\documentclass{article}

 \usepackage[preprint]{neurips_2026}

\usepackage{tipa}
\usepackage{todonotes}
\usepackage{multirow}
\usepackage{textcomp}
\usepackage{colortbl}
\usepackage{mathtools}
\usepackage{amssymb}
\usepackage{amsthm}
\usepackage{tabu}
\usepackage{siunitx}
\usepackage{enumitem}
\usepackage{rotating}
\usepackage{amsmath}
\usepackage{wrapfig}
\usepackage{caption}

\definecolor{third}{rgb}{1, 1, 0.7}
\definecolor{second}{rgb}{1, 0.85, 0.7}
\definecolor{first}{rgb}{1, 0.7, 0.7}

\definecolor{lightgreen}{rgb}{0.9, 1.0, 0.9}
\definecolor{lightred}{rgb}{1.0, 0.9, 0.9}
\newcommand{\btr}[1]{\cellcolor{lightgreen}#1}
\newcommand{\wrs}[1]{\cellcolor{lightred}#1}
\usepackage[utf8]{inputenc} 
\usepackage[T1]{fontenc}    
\usepackage{hyperref}       
\usepackage{url}            
\usepackage{booktabs}       
\usepackage{amsfonts}       
\usepackage{nicefrac}       
\usepackage{microtype}      
\usepackage{xcolor}         

\title{Towards Complementary Keypoint Detection via Mixture of Detectors}

\author{%
  Xiaoyong Lu \\
  Southeast University\\
  \texttt{luxiaoyong@seu.edu.cn} \\
  \And
  Guobao Xiao \\
  Tongji University \\
  \texttt{gbx@tongji.edu.cn} \\
  \And
  Dong Liang \\
  Nanjing University of Aeronautics and Astronautics \\
  \texttt{liangdong@nuaa.edu.cn} \\
  \And
  Songlin Du\thanks{Corresponding author.} \\
  Southeast University \\
  \texttt{sdu@seu.edu.cn} \\
}

\begin{document}

\maketitle

\begin{abstract}
Keypoint detection plays a central role in geometric vision pipelines. 
Existing detectors exhibit complementary strengths across different scene structures and keypoint budgets, 
and reliance on a single detector often results in suboptimal keypoint distributions. 
Motivated by this observation, we propose \textbf{Mixture of Detectors (MoD)} for dense matching pipelines, 
a unified framework for complementary keypoint detection through the adaptive composition of heterogeneous detectors. 
MoD first distills heterogeneous detectors into a unified model, 
allowing diverse detection maps to be predicted efficiently within a single forward pass. 
Detector selection is subsequently formulated as a patch-wise, budget-conditioned decision process, 
in which a lightweight router assigns the most suitable detector to each local region.
To avoid heuristic supervision, the router is trained with task-driven rewards derived from downstream matching quality, 
jointly capturing both keypoint quantity and precision. 
Extensive experiments on MegaDepth, ScanNet, and HPatches demonstrate that MoD consistently outperforms strong baselines 
across a wide range of keypoint budgets, demonstrating its effectiveness in 
improving keypoint distribution and downstream geometric performance.
\end{abstract}
    
\section{Introduction}
\label{sec:intro}
Keypoint detection aims to identify distinctive image structures and constitutes a fundamental component of geometric vision pipelines, 
including Structure-from-Motion (SfM) \cite{sfm}, Simultaneous Localization and Mapping (SLAM) \cite{slam}, and image matching \cite{sarlin20superglue}. 
In classical sparse matching pipelines, two sets of keypoints are detected and subsequently matched, 
and the quality of matching relies heavily on repeatability under varying illumination and viewpoint conditions. 
Recent dense matching approaches \cite{edstedt2024roma, rcm-plus} have partially relaxed this requirement by 
detecting keypoints in one image and directly establishing correspondences in another,
shifting the focus from strict repeatability to the spatial distribution of detected keypoints. 
As a result, how to obtain an effective and adaptive keypoint distribution becomes increasingly important for downstream performance.

In practice, different detectors exhibit complementary behaviors across scene structures and keypoint budgets. 
This variation stems from their distinct design choices, including normalization strategies (\emph{e.g.}, local or global Softmax versus independent Sigmoid), 
training objectives (\emph{e.g.}, supervised learning, self-supervised learning, or reinforcement learning), 
and the types of geometric primitives they capture (\emph{e.g.}, points, lines, and edges). 
\begin{wrapfigure}{r}{0.5\linewidth}
    \centering
    \includegraphics[width=\linewidth]{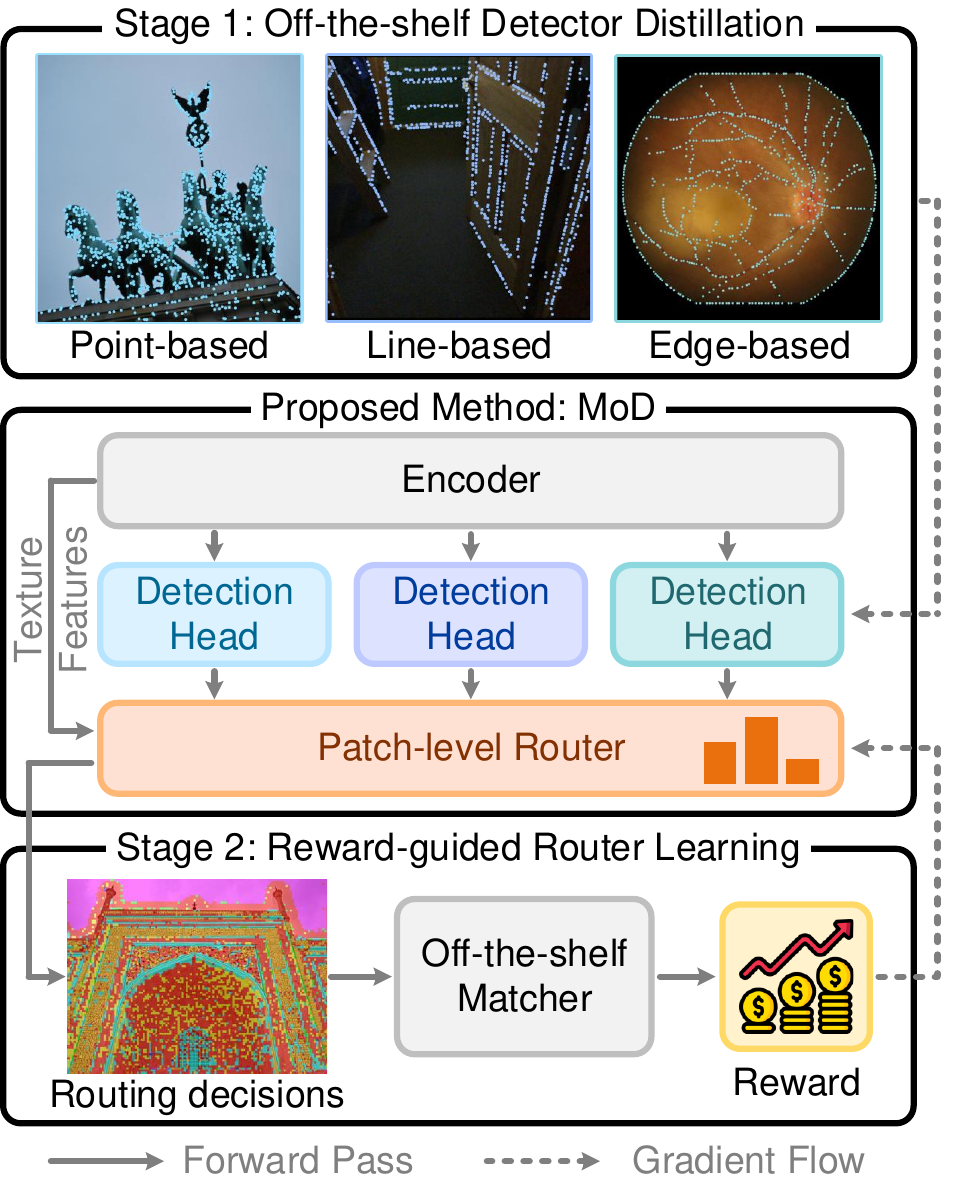}
    \caption{
    Different detectors exhibit complementary behaviors across varying image structures and keypoint budgets.
    MoD leverages this by first distilling heterogeneous detectors into a unified model for joint prediction (Stage~1),
    and then learning a reward-guided router that selects detectors based on local texture and the keypoint budget (Stage~2).
    }
    \label{fig:fig1}
    \vspace{-1em}
\end{wrapfigure}
These differences lead to substantially different detection patterns: some detectors favor sparse and highly reliable keypoints, 
while others provide denser coverage over structured or low-texture regions. 
Moreover, their relative performance varies significantly with the target number of keypoints, \emph{i.e.}, the keypoint budget. 
Consequently, no single detection mechanism performs consistently well across all spatial regions and budget settings.

Prior work such as RCM+ \cite{rcm-plus} shows that combining multiple detectors improves performance due to complementary distributions. 
However, such approaches rely on simple concatenation or heuristic fusion, 
which does not explicitly model detector selection and cannot fully exploit complementary strengths. 
In particular, the optimal detector varies across images and depends on \textbf{local textures} and the target \textbf{keypoint budget}.
This leads to a more general formulation, 
where keypoint detection selects suitable detectors across spatial locations under a given budget. 
Such a formulation introduces several challenges:  
(i) running all candidate detectors is computationally expensive, 
(ii) detector selection must jointly account for local textures and global keypoint budget, 
and (iii) no direct supervision is available for detector selection.

To address these challenges, we propose \textbf{Mixture of Detectors (MoD)},  
a unified framework for adaptive detector composition. 
Motivated by the Mixture-of-Experts (MoE) paradigm \cite{shazeer2017outrageously}, 
we treat point-, line-, and edge-based detectors as specialized experts, 
and formulate selection as a routing problem over these experts. 
(i) To reduce the cost of multiple detectors, 
MoD distills heterogeneous detectors into a single model that predicts diverse detection maps in a forward pass, 
preserving complementary behaviors while maintaining efficiency. 
(ii) Detector selection is formulated as a patch-wise, budget-conditioned process. 
A lightweight router assigns detectors to regions based on local texture, 
while a budget-aware mechanism introduces a global prior that modulates preference by desired keypoint density.
(iii) To address the lack of supervision, 
detector selection is guided by task-driven rewards from downstream matching performance, 
providing patch-level supervision that captures keypoint quantity and precision. 

To summarize, this paper makes the following contributions:
\textbf{(1)} We present a new perspective on keypoint detection by modeling the complementary behaviors of heterogeneous detectors, 
and introduce a detector selection problem conditioned on spatial regions and keypoint budgets.
\textbf{(2)} We introduce a unified detector that distills heterogeneous detectors into a shared model, 
enabling efficient prediction of diverse detection maps in a single forward pass.
\textbf{(3)} We propose a patch-wise, budget-aware router that adaptively selects detectors based on local texture and global keypoint budget, 
trained with task-driven rewards that jointly capture keypoint quantity and precision.
\textbf{(4)} Extensive experiments on MegaDepth, ScanNet, and HPatches show that MoD consistently improves downstream geometric estimation 
across a wide range of keypoint budgets while maintaining competitive efficiency.

\section{Related Work}
\label{sec:related}

\paragraph{Detection.}
Early keypoint detection methods relied on handcrafted designs, among which SIFT \cite{liu2010sift} is representative. 
These methods detect salient structures using manually designed operators, 
achieving reasonable repeatability under moderate variations. 
Recent learning-based approaches have advanced keypoint detection. 
Methods such as SuperPoint \cite{detone2018superpoint}, D2-Net \cite{dusmanu2019d2}, DISK \cite{tyszkiewicz2020disk}, 
and ALIKED \cite{zhao2023aliked} use deep neural networks to jointly learn detection and description from data,
improving robustness across applications. 
Furthermore, DeDoDe \cite{edstedt2024dedode,edstedt2024dedodev2} and DaD \cite{edstedt2025dad} adopt a decoupled detection-description paradigm, 
motivated by inconsistent optimization objectives between detection and description. 
Joint optimization can lead to suboptimal solutions and restrict flexibility of keypoint distributions. 
Following this line, MoD adopts a fully decoupled detection formulation.
Beyond point-based detection, line segment detection methods such as LSD \cite{lsd}, SOLD$^2$ \cite{SOLD2}, 
and DeepLSD \cite{deeplsd} capture structured geometric primitives, particularly in man-made environments with prevalent linear structures. 
Similarly, edge detection methods, including Canny \cite{canny}, EDTER \cite{pu2022edter}, and MuGE \cite{muge}, 
identify object boundaries and fine-grained contours often missed by point-based detectors. 
These detection methods provide complementary information to point-based detection, enriching spatial coverage.
However, the complementarity among different detection methods remains largely under-explored. 

\paragraph{Mixture of Experts.}
Mixture of Experts (MoE) \cite{shazeer2017outrageously,riquelme2021scaling,moe,adamv} is a widely adopted paradigm that 
improves model capacity through experts and a routing mechanism. 
In most works, particularly in large-scale language and vision models, experts correspond to different feed-forward networks with the same architecture, 
and the router activates a subset of them for each input token.
Such designs enhance representation learning and computational efficiency via conditional computation.
In this work, different types of detectors, including point-, line-, and edge-based methods, 
are treated as experts rather than multiple networks with similar functions.
These detectors capture distinct geometric structures and exhibit complementary behaviors across scenes, 
improving keypoint distribution and downstream performance in dense matching pipelines.

\section{Method}
\label{sec:method}
\begin{figure*}
    \centering
    \includegraphics[width=\linewidth]{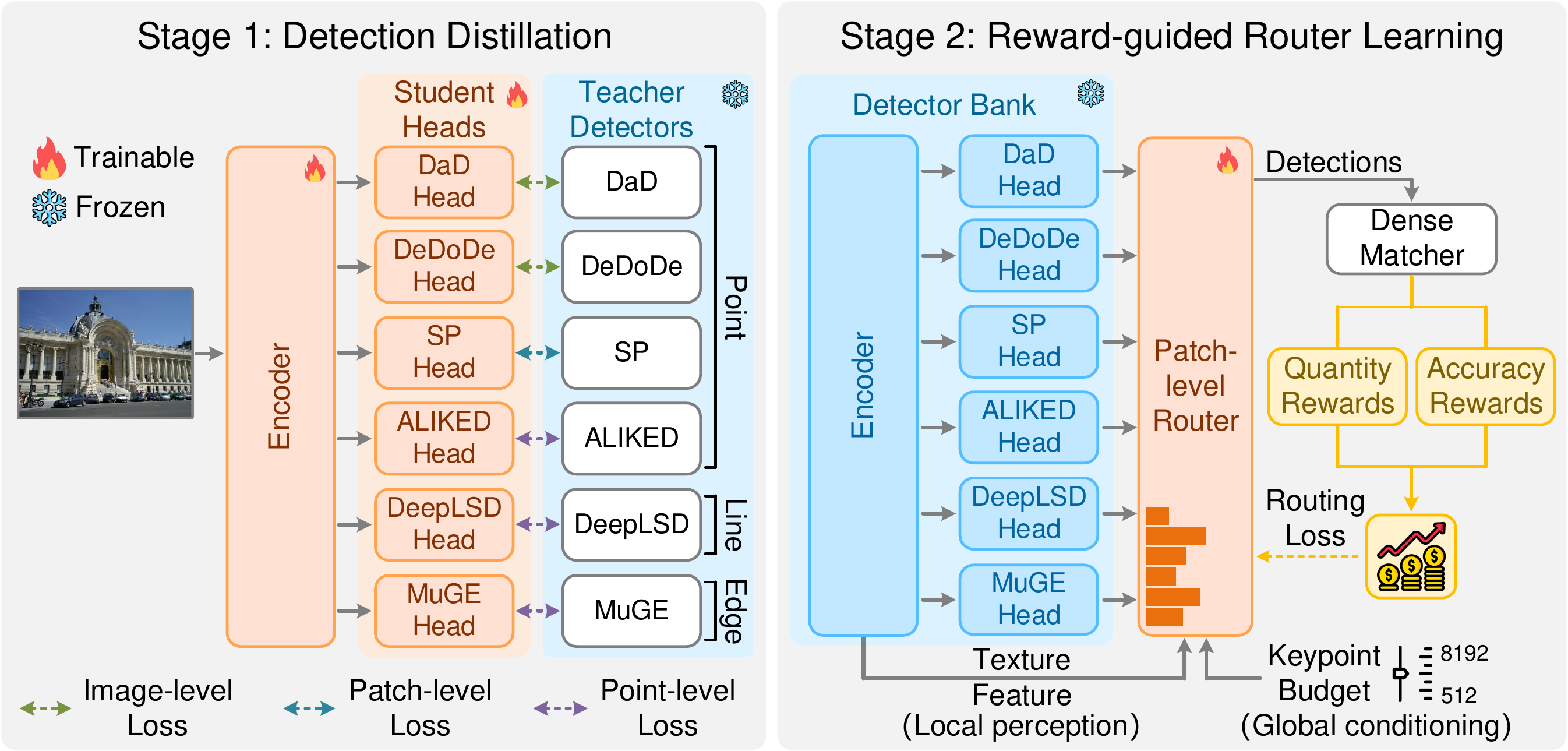}
    \caption{\textbf{Overview of the proposed method.} 
    In stage 1, we distill multiple teacher detectors into student heads. 
    In stage 2, we train a patch-level router with reward-guided supervision to select detectors based on local texture and keypoint budget.}
    \label{fig:structure}
\end{figure*}

As illustrated in Figure \ref{fig:structure},  
MoD employs a shared encoder to extract features, followed by multiple detection heads that produce detection maps. 
In Stage 1, distillation from six state-of-the-art detectors enables MoD to generate six maps in a single forward pass. 
In Stage 2, the encoder and heads are frozen, and a router uses texture features and keypoint budget to assign the most suitable detector to each $8\times8$ patch. 
Routing decisions are evaluated with an off-the-shelf matcher \cite{edstedt2024roma}, 
which provides rewards based on keypoint quantity and precision to supervise training.

\subsection{Stage 1: Detection Distillation}
\subsubsection{Feature Extraction}
We adopt a VGG \cite{simonyan2015vgg} network as the encoder to extract multi-scale dense feature maps from the input image $I$. 
VGG provides rich local texture features, essential for keypoint detection and subsequent routing. 
Specifically, features are collected from different backbone stages to form a feature pyramid $\{F_{1}, F_{1/2}, F_{1/4}, F_{1/8}\}$, 
where each subscript denotes the spatial resolution relative to $I$.
The unified encoder aggregates knowledge from multiple teachers during distillation, 
yielding more robust and generalizable texture representations. 
As a result, \emph{the student may even outperform individual teachers}.

\subsubsection{Detection Head} 
Because detectors produce detection probabilities via different mechanisms, 
we introduce two types of detection heads: \textbf{image-level} and \textbf{patch-level}. 
Specifically, ALIKED \cite{zhao2023aliked}, DeDoDe \cite{edstedt2024dedodev2}, DaD \cite{edstedt2025dad}, 
DeepLSD \cite{deeplsd}, and MuGE \cite{muge} generate keypoint probabilities at the image level, 
whereas SuperPoint \cite{detone2018superpoint} predicts probabilities at the patch level.

\textbf{Image-level head.}
For detectors operating on full-resolution maps, 
we adopt depthwise convolution refiners from DKM \cite{edstedt2023dkm}. 
The prediction follows a coarse-to-fine refinement across $\{F_{1/8}, F_{1/4}, F_{1/2}, F_{1}\}$. 
At each scale, a refiner takes the encoder feature and predicts a residual added to upsampled logits from the previous scale. 
This progressive refinement recovers fine spatial details while preserving global consistency.
After obtaining the final logits, different normalization functions are applied to align with teacher behaviors.
For DeDoDe and DaD, we apply Softmax over all pixels. 
For ALIKED, DeepLSD, and MuGE, we apply Sigmoid independently per pixel.

\textbf{Patch-level head.}
For SuperPoint, which predicts keypoints at the patch level, we introduce a head on $F_{1/8}$. 
Specifically, a VGG-style block predicts a $65$-channel logit map, 
where $64$ channels correspond to the $8\times8$ locations within each patch and one channel represents a \emph{dustbin} class, \emph{i.e.}, a no-keypoint class. 
Softmax is applied over the $65$ channels, then the dustbin channel is discarded. 
The remaining probabilities are reshaped to recover a dense detection map at the original resolution.

\subsubsection{Learning Objective}
To faithfully distill heterogeneous detectors, 
we adopt different objectives according to how each method generates detection probabilities.
\textbf{(1) Image-level Supervision.}
For DeDoDe and DaD, which generate probabilities via spatial Softmax over image-level logits, 
we use image-level Kullback-Leibler (KL) divergence to align the predicted distributions. 
\textbf{(2) Patch-level Supervision.}
For SuperPoint, which predicts probabilities in each $8\times8$ patch,
we compute KL divergence within each patch to align the predicted distributions.
\textbf{(3) Point-level Supervision.}
For ALIKED, DeepLSD, and MuGE, which model probabilities independently per pixel using Sigmoid, 
we regress the logit maps with an $\ell_2$ loss.
The overall objective sums the losses over all detectors.

\subsection{Stage 2: Reward-guided Router Learning} 
\subsubsection{Router}
\label{sec:router}
\begin{wrapfigure}{r}{0.5\linewidth}
    \vspace{-0.5em}
    \centering
    \includegraphics[width=\linewidth]{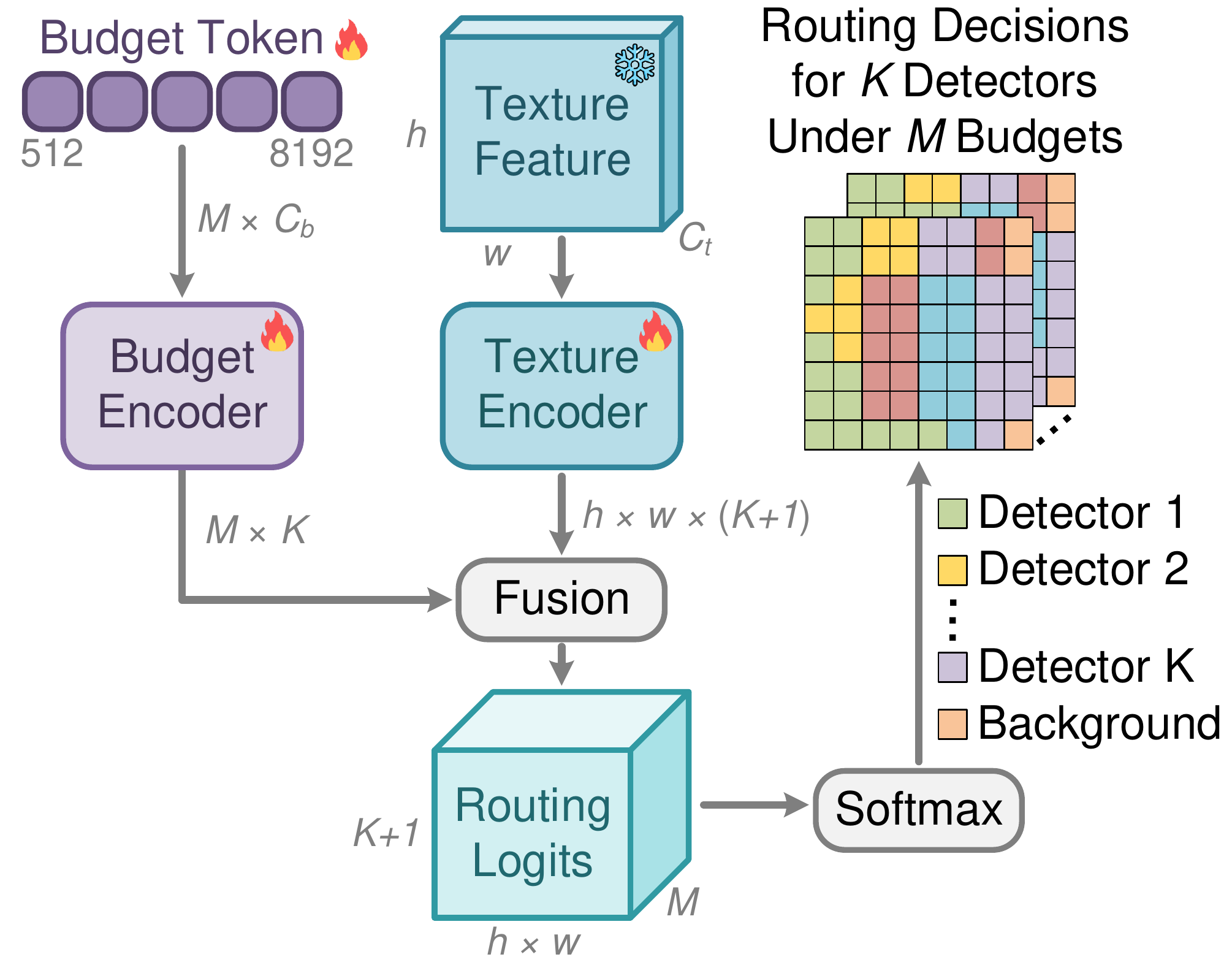}
    \caption{\textbf{Detector router conditioned on keypoint budget and local texture.}
    }
    \label{fig:numbias2}
    \vspace{-1em}
\end{wrapfigure}

In Stage 2, a router is trained with reward-guided supervision to select the most suitable detector per patch based on local texture and global keypoint budget.
As illustrated in Figure \ref{fig:numbias2}, the router has two branches, 
a \textbf{budget branch} and a \textbf{texture branch}, whose outputs are fused for final routing decisions.

\textbf{Texture branch.}  
The texture branch captures local patterns to determine the appropriate detector for each region. 
It takes the frozen encoder feature at $1/8$ resolution and processes it through a CNN-based encoder to produce a spatial feature map. 
This branch models local structures such as corners, edges, and low-texture regions, 
enabling the router to exploit spatial heterogeneity. 
Let $h = H/8$ and $w = W/8$. 
The output, denoted by $\hat{R}\in\mathbb{R}^{(K+1) \times h \times w}$, provides patch-wise preferences over detectors. 
Here, $K$ is the number of detectors, and the additional class represents a background option where no detector is selected.

\textbf{Budget branch.}  
In addition to texture, detector selection depends on the desired number of keypoints. 
We introduce a budget branch that encodes the target keypoint budget into a learnable representation. 
Specifically, $M=5$ budget tokens, corresponding to keypoint budgets $\{512, 1024, 2048, 4096, 8192\}$, are learned during training.
Budget tokens are processed through a MLP-based encoder to produce a global bias $b\in\mathbb{R}^{K\times M}$.
In addition to these five budgets, other budgets can be handled by interpolating or extrapolating the budget tokens, 
\emph{e.g.}, linearly interpolating the $1024$ and $2048$ tokens to obtain the detector bias for a budget of $1500$.

\textbf{Fusion and routing decision.}  
The outputs of the two branches are combined to produce routing logits. 
For each budget level $m \in \{1,\dots,M\}$, the bias $b^m \in \mathbb{R}^{K}$ is added to the first $K$ channels of $\hat{R}$.
The routing logits are denoted by $R\in\mathbb{R}^{(K+1)\times M \times h \times w}$. 
The budget bias is broadcast to all locations and acts as a global prior modulating detector preference by keypoint density. 
The fused logits are passed through a Softmax along the detector dimension to yield selection probabilities for each $8\times8$ patch, 
denoted by $P \in \mathbb{R}^{(K+1) \times M \times h \times w}$.

\begin{wrapfigure}{r}{0.6\linewidth}
    \centering
    \vspace{-1em}
    \includegraphics[width=\linewidth]{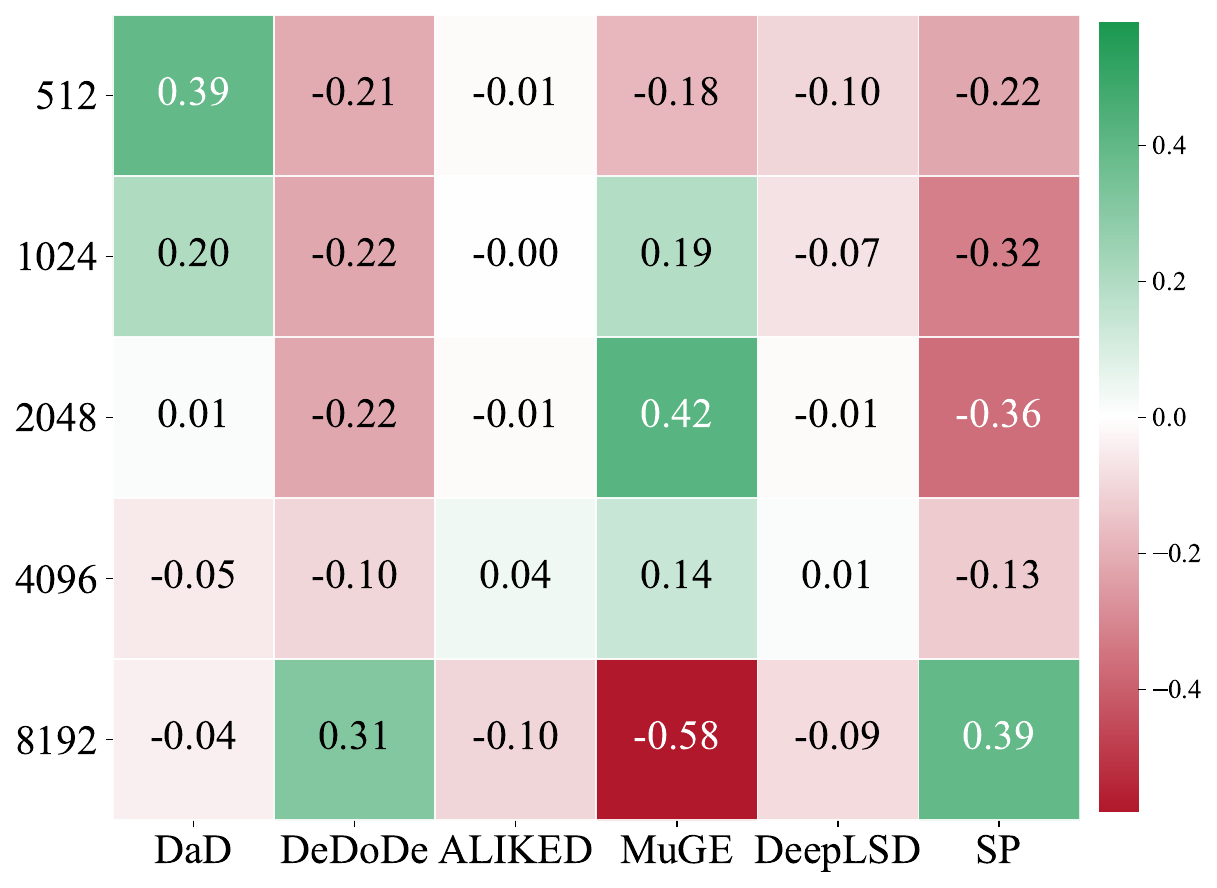}
    \caption{\textbf{Learned biases of different detectors under varying keypoint budgets.} 
    The vertical axis denotes the budgets, and the horizontal axis corresponds to the detectors.}
    \label{fig:bias}
    \vspace{-1em}
\end{wrapfigure}

The router can be interpreted as a patch-wise policy determining how the final detection map is composed from distilled heads. 
By jointly modeling texture and budget, the router enables flexible, context-aware detector selection, 
allowing detectors to specialize across local regions and keypoint densities.

\subsubsection{Reward}
We construct a patch-level reward from matching quality of keypoints predicted by different detection heads under varying budgets.
We extract keypoints from all $K$ detection heads under $M$ budgets in source image,
and use an off-the-shelf matcher \cite{romav2} to find correspondences in target image.
Let $x$ denote keypoints in source image, $x_p$ their predicted matched positions, and $x_{gt}$ the ground-truth matched positions.
We compute the error $e$ for each match as the Euclidean distance between $x_p$ and $x_{gt}$.
Given threshold $\tau$, a keypoint is an inlier if $e<\tau$, yielding two rewards.

\textbf{Quantity reward.}
The first reward $r^{\text{qty}}_{p,k}$ evaluates the ability of each detector to produce a larger number of reliable keypoints within a patch,
where $p$ indexes the patch and $k \in {1,\dots,K}$ denotes the detector class.
We compute the number of inliers for each detector within each patch.
Detectors are then ranked according to the inlier counts within each patch, 
and a reward of $1$ is assigned to the top-k detector, while the remaining detectors receive $0$. 

\textbf{Precision reward.}
The second reward $r^{\text{prec}}_{p,k}$ measures the precision of keypoints within a patch. 
For each detector, we compute the average error of its inlier matches within each patch, since detectors whose keypoints are more precisely localized on distinctive textures tend to produce more accurate correspondences in the target image. 
Detectors are then ranked by this average error in each patch, and a reward of $1$ is assigned to the top-k detector, while the remaining detectors receive $0$.

The final reward combines both terms as a weighted sum:
\begin{align}
    & r_{p,k} = \alpha r^{\text{qty}}_{p,k} + (1-\alpha) r^{\text{prec}}_{p,k},
\end{align}
where $\alpha$ controls the trade-off between keypoint quantity and precision. 
Since optimal detector assignment depends on the target keypoint budget, 
rewards are computed per budget $m \in \{512, 1024, 2048, 4096, 8192\}$ using the top-$m$ keypoints of each detector and normalized across detectors. 
Specifically, for each patch and budget, the reward of each detector is divided by the sum of rewards over all detectors. 
Stacking all budgets yields $r \in \mathbb{R}^{K \times M \times h \times w}$, encoding detector preferences conditioned on local texture and keypoint budget.

\subsubsection{Learning Objective}
We train the router with a \emph{routing loss} that favors high-reward detectors and a \emph{background loss} that suppresses assignments in invalid regions.

\textbf{Routing loss.}
Given the routing probabilities $P \in \mathbb{R}^{(K+1) \times M \times h \times w}$ defined in Section \ref{sec:router}, 
we exclude its background class and compute a reward-guided cross-entropy loss
\begin{align}
    & \mathcal{L}_{\text{router}} =
- \frac{1}{MKhw}
\sum_{m,k,p}
r^m_{k,p}\log P^m_{k,p},
\end{align}
encouraging higher probability for detectors with better rewards.

\textbf{Background loss.}
We derive background supervision from the depth maps of MegaDepth \cite{li2018megadepth}, where unstable regions such as sky are assigned zero depth.  
We then compute the cross-entropy loss between the predicted background probabilities and background ground truth.

The final objective for stage 2 is defined as the sum of the two losses:
\begin{align}
    & \mathcal{L}
=
\mathcal{L}_{\text{router}}
+
\mathcal{L}_{\text{background}}.
\end{align}

\section{Experiments}
\label{sec:results}

\subsection{Implementation Details} 
\label{sec:details}
In stage 1, the encoder is initialized with ImageNet pretraining 
and jointly optimized with all detection heads using AdamW~\cite{adamw}. 
The learning rate is $1\times10^{-4}$ with weight decay $1\times10^{-5}$, 
and the model is trained for $200{,}000$ iterations with batch size $4$. 
In stage 2, the router is trained with AdamW using a learning rate of $2\times10^{-4}$ and weight decay $1\times10^{-5}$, for $20{,}000$ iterations with batch size $4$. 
Rewards are assigned to the top-2 detectors for both quantity and precision rewards. 
The error threshold is $1$ pixel, and the trade-off parameter in the reward is $\alpha = 0.8$. 
All models are trained on a NVIDIA 4090 GPU. 
Following DaD \cite{edstedt2025dad}, the longer image side is resized to $1024$ during evaluation.
We use RoMa v2 \cite{romav2} as the matcher during training, and evaluate with both RoMa \cite{edstedt2024roma} 
and RCM+ \cite{rcm-plus} to assess cross-matcher generalization and avoid overfitting to a single matcher.

\subsection{Pose and Homography Estimation}
\label{sec:exp}

\textbf{Datasets.}  
MegaDepth~\cite{li2018megadepth} contains one million Internet images spanning $196$ outdoor scenes, with sparse 3D reconstructions from COLMAP and depth maps from multi-view stereo. 
We evaluate on 1,500 image pairs from the Sacre Coeur and St.~Peter's Square scenes following LoFTR~\cite{sun2021loftr}. 
ScanNet~\cite{dai2017scannet} contains $1613$ monocular sequences with ground-truth poses and depth maps. 
We evaluate on 1,500 image pairs from SuperGlue~\cite{sarlin20superglue}. 
The HPatches dataset~\cite{balntas2017hpatches} includes $52$ illumination and $56$ viewpoint sequences. 
Following DaD \cite{edstedt2025dad}, we evaluate on the $56$ viewpoint sequences with geometric transformations.

\noindent\textbf{Metrics.}  
For MegaDepth and ScanNet, we report the AUC of pose error at $5^\circ$ under five keypoint budgets: $512$, $1024$, $2048$, $4096$, and $8192$. 
For HPatches, we report the AUC of end-point error at $3$ pixels under the same five keypoint budgets.

\begin{table*}[t]
    \centering
    \caption{\textbf{Pose Estimation and Homography Estimation using RoMa as the matcher.} 
    Performance is measured as the AUC@5$^{\circ}$ of the pose error and the AUC@3 pixels of the end-point error.
    MoD$^\dagger$ denotes the student detection head corresponding to the teacher detector.}
    \small 
    \setlength{\tabcolsep}{3pt}
    \begin{tabular}{l ccccccccccccccccc}
    \toprule
        \multirow{2}{*}{Method} & \multicolumn{5}{c}{MegaDepth} & \multicolumn{5}{c}{ScanNet} & \multicolumn{5}{c}{HPatches} \\
        \cmidrule(lr){2-6}
        \cmidrule(lr){7-11}
        \cmidrule(lr){12-16}
        & 512 & 1024 & 2048 & 4096 & 8192 & 512 & 1024 & 2048 & 4096 & 8192 & 512 & 1024 & 2048 & 4096 & 8192 \\
        \midrule
        SP & 64.6& 65.4 & 65.7	& 65.4 & 64.3 & 28.4 & 28.7 & 28.6 & 28.9 & 28.6 & 58.6 & 59.4 & 59.1 & 59.4 & 58.9 \\
        MoD$^\dagger$ & \btr 64.6 & \btr 65.4 & \wrs 65.4 & \btr 65.6 & \btr 65.0 & \btr 28.5 & \btr 28.9 & \btr 28.6 & \btr 29.0 & \btr 29.3 & \btr 60.4 & \btr 61.0 & \btr 60.6 & \btr 60.8 & \btr 60.7 \\
        \midrule
        ALIKED & 65.8 & 65.7 & 65.6 & 65.9 & 65.9 & 25.7 & 26.4 & 27.2 & 27.7 & 27.9 & 56.8 & 57.6 & 57.9	& 57.4 & 57.4 \\
        MoD$^\dagger$ & \btr 66.3 & \btr 66.6 & \btr 66.4 & \btr 66.7 & \wrs 65.3 & \btr 26.3 & \btr 27.1 & \btr 27.3 & \btr 27.8 & \btr 28.1 & \btr 59.9 & \btr 59.9 & \btr 59.7 & \btr 59.6 & \btr 59.6 \\
        \midrule
        DeDoDe & 63.3	& 64.7 & 65.6 & 65.7 & 66.4 & 27.0 & 27.7 & 28.2 & 28.6 & 28.7 & 58.1 & 59.7 & 61.1 & 61.9 & 62.1 \\
        MoD$^\dagger$ & \btr 63.6 & \btr 65.3 & \btr 65.8 & \btr 66.1 & \btr 66.4 & \btr 27.1 & \btr 27.8 & \btr 28.3 & \btr 28.6 & \btr 28.9 & \btr 58.8 & \btr 60.2 & \btr 61.3 & \btr 61.9 & \btr 62.1\\
        \midrule
        DaD & 66.3 & 67.2 & 67.6 & 67.8 & 66.0 & 28.7 & 29.0 & 29.1 & 29.1 & 29.0 & 60.5 & 61.1 & 61.0 & 60.8 & 60.6 \\
        MoD$^\dagger$ & \wrs 66.2 & \btr 67.5 & \btr 67.7 & \btr 67.8 & \btr 66.1 & \btr 29.0 & \btr 29.4 & \btr 29.7 & \btr 29.5 & \btr 29.1 & \btr 60.7 & \btr 61.3 & \btr 61.1 & \btr 61.0 & \btr 60.9 \\
        \midrule
        DeepLSD & 57.3 & 60.8 & 63.1 & 65.0 & 63.7 & 18.2 & 22.5 & 26.4 & 28.2 & 27.5 & 48.6 & 52.3 & 55.6 & 59.2 & 59.3 \\
        MoD$^\dagger$ & \btr 58.0 & \btr 61.8 & \btr 64.0 & \btr 65.1 & \btr 64.3 & \wrs 17.9 & \btr 23.1 & \btr 26.6 & \btr 28.8 & \btr 28.2 & \btr 52.1 & \btr 55.0 & \btr 57.5 & \btr 60.2 & \btr 60.7\\
        \midrule
        MuGE & 59.6 & 63.1 & 64.1 & 63.5 & 63.5 & 19.7 & 23.0 & 25.8 & 27.4 & 28.7 & 51.6 & 53.5 & 56.6 & 58.3 & 59.6 \\
        MoD$^\dagger$ & \btr 62.2 & \btr 63.8 & \btr 64.2 & \btr 64.6 & \wrs 63.0 & \btr 21.5 & \btr 25.4 & \btr 28.5 & \btr 28.6 & \btr 28.7 & \btr 53.7 & \btr 56.9 & \btr 59.9 & \btr 60.9 & \btr 59.9\\
        \midrule
        MoD & \btr 66.3 & \btr 67.5 & \btr 68.3 & \btr 68.4 & \btr 68.5 & \wrs 28.8 & \btr 29.9 & \btr 30.1 & \btr 30.3 & \btr 29.3 & \btr 60.7 & \btr 62.4 & \btr 62.5 & \btr 62.6 & \btr 62.4 \\
         \bottomrule
    \end{tabular}
    \label{tab:all_1}
    \vspace{-1.5em}
\end{table*}

\noindent\textbf{Results.}  
As shown in Table \ref{tab:all_1}, MoD$^\dagger$, which denotes the student detection head, outperforms the original detectors in most settings, indicating effective multi-teacher distillation. 
On MegaDepth, gains are consistent for ALIKED, DeDoDe, and DeepLSD, with only mild degradations such as MuGE at $8192$.
Full MoD with router achieves the best performance at all budgets, reaching $66.3$, $67.5$, $68.3$, $68.4$, and $68.5$. 
On ScanNet, MoD achieves the best performance at four budgets and, at $512$, remains above the original DaD despite being lower than its student head. 
The larger margin suggests stronger benefits in indoor scenes. 
On HPatches, MoD$^\dagger$ improves in all cases, and full MoD achieves the highest AUC at all budgets, 
reaching $60.7$, $62.4$, $62.5$, $62.6$, and $62.4$, consistently outperforming DaD. 
Across datasets, detector-wise trends are consistent, with DeDoDe improving at high budgets and DaD remaining competitive at low budgets, supporting budget-aware adaptive selection.
Additional evaluations using RCM+ \cite{rcm-plus} as the matcher are provided in the Appendix \ref{sec:rcm}.

\subsection{Ablation Study}
As shown in Table~\ref{tab:ablation}, we analyze the contribution of key design components in MoD, 
including the budget awareness, background modeling, routing strategy, and reward formulation.

\begin{wraptable}{r}{0.50\linewidth}
    \vspace{-1em}
    \centering
    \small
    \setlength{\tabcolsep}{4pt}
    \caption{\textbf{Ablation Study on HPatches.} Performance is measured by AUC@3 pixels of the end-point error.}
    \begin{tabular}{l ccccc}
    \toprule
        \multirow{2}{*}{Method} & \multicolumn{5}{c}{Homography Estimation AUC $\uparrow$} \\
        \cmidrule(lr){2-6}
        & 512 & 1024 & 2048 & 4096 & 8192 \\
        \midrule
        \emph{w/o} budget bias & 59.2 & 61.0 & 61.4 & 61.7 & 61.7 \\
        \emph{w/o} background & 59.5 & 61.1 & 61.9 & 61.6 & 62.4 \\
        \midrule
        avg. det. map & 59.9 & 60.3 & 60.8 & 61.0 & 61.2 \\
        max det. map & 60.0 & 60.4 & 61.0 & 60.6 & 60.8 \\
        global routing & 60.2 & 60.5 & 61.2 & 61.2 & 61.5 \\
        soft routing  & 60.2 & 60.8 & 61.3 & 61.5 & 61.4 \\
        \midrule
        $\alpha=0$ & 59.9 & 61.5 & 62.0 & 62.6 & 62.3 \\
        $\alpha=1$ & 59.9 & 62.0 & 62.1 & 62.4 & 62.3 \\
        $\alpha=0.5$ & 60.0 & 62.0 & 62.1 & 62.5 & 62.4 \\
        \midrule
        MoD & \textbf{60.7} & \textbf{62.4} & \textbf{62.5} & \textbf{62.6} & \textbf{62.4} \\
         \bottomrule
    \end{tabular}
    \label{tab:ablation}
    \vspace{-2em}
\end{wraptable}

\begin{figure*}
    \centering
    \includegraphics[width=\linewidth]{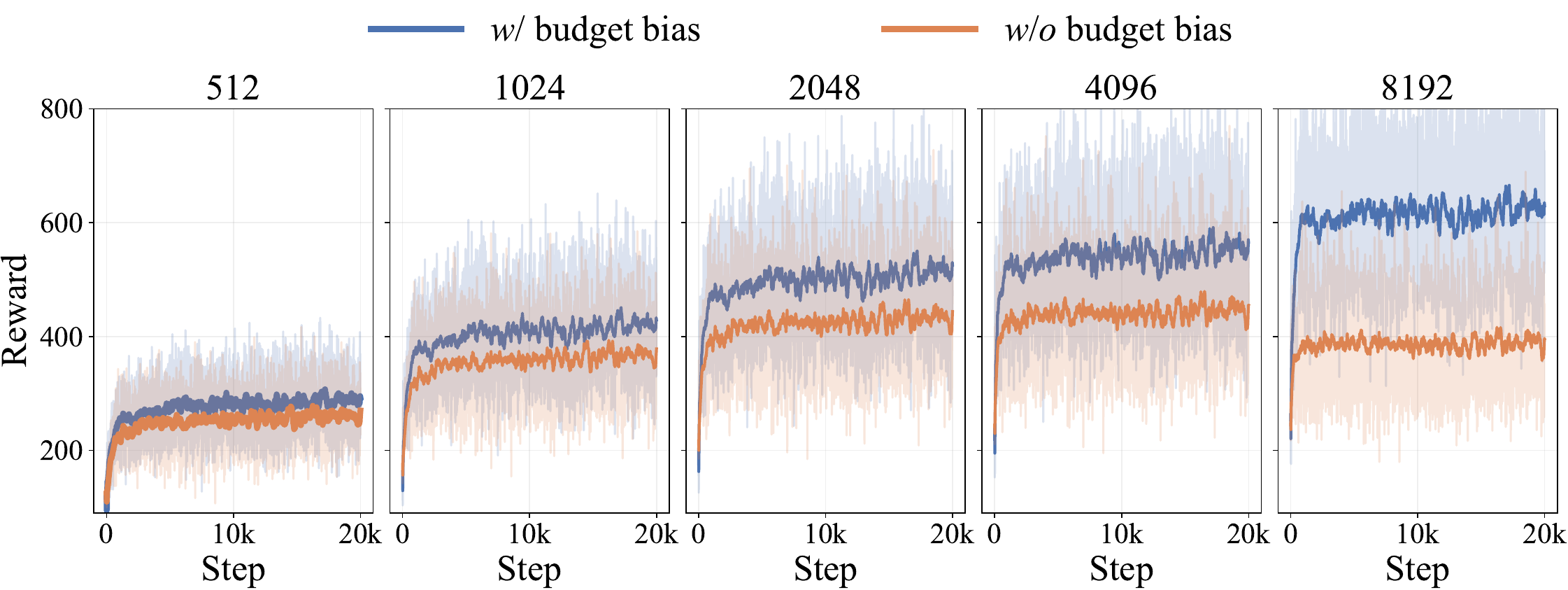}
    \vspace{-1.5em}
    \caption{\textbf{Reward curves under different keypoint budgets with and without budget bias.}
    MoD without budget bias exhibits conflicting detector preferences across budgets.
    }
    \label{fig:reward}
    \vspace{-0.5em}
\end{figure*}

\textbf{Budget Awareness and Background Modeling.}   
Removing the budget bias significantly degrades performance, particularly at higher keypoint budgets. 
This suggests that conditioning the router on the target keypoint count provides a useful global prior. 
A similar trend appears in Figure \ref{fig:reward}, which shows the training rewards of the router. 
Without budget awareness, the reward at 8192 is lower than at 4096, indicating conflicting routing decisions across budgets. 
In contrast, introducing budget awareness improves rewards across all settings, with larger gains at higher budgets. 

Removing the background class also degrades performance at several budgets. 
This suggests that modeling non-matchable regions helps the router avoid low-quality areas, 
improving the quality and stability of routing decisions.

\begin{figure*}
    \centering
    \includegraphics[width=\linewidth]{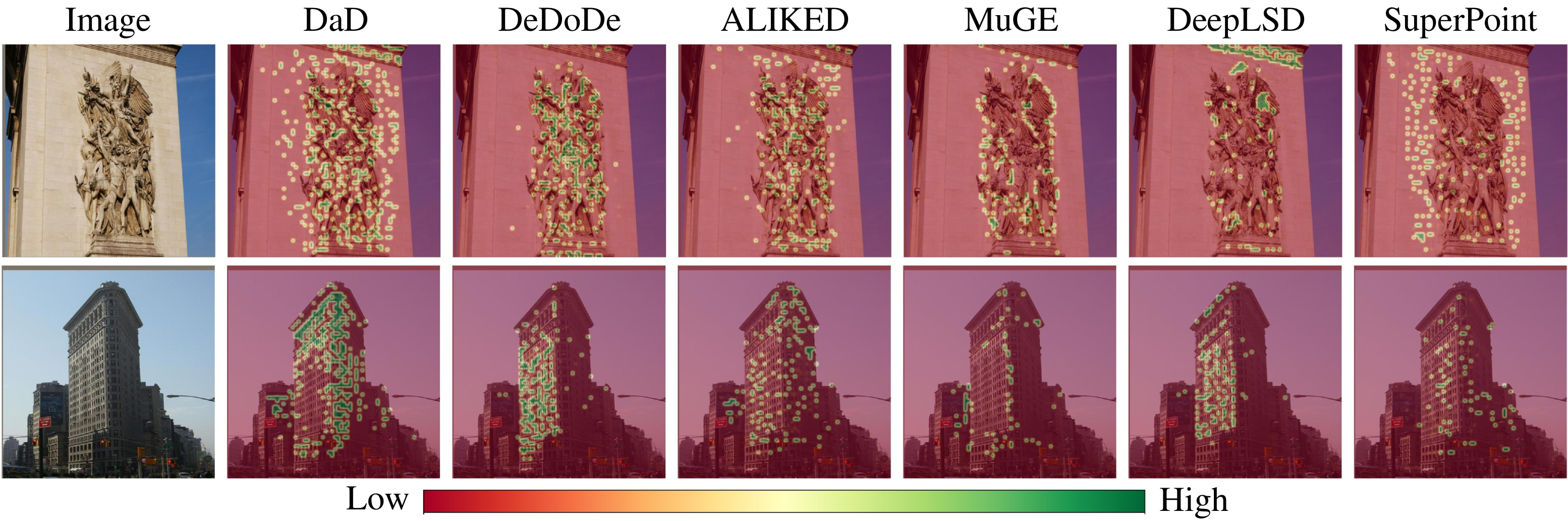}
    \vspace{-1.5em}
    \caption{\textbf{Reward distributions for different detection methods} (best viewed when zoomed in).}
    \label{fig:router spv}
    \vspace{-0.5em}
\end{figure*}

\textbf{Routing Strategy.}
We investigate alternative routing and fusion strategies based on the six distilled detection maps.
The avg.\ det.\ map and the max.\ det.\ map denote non-learnable fusion schemes that combine detection maps through pixel-wise averaging and maximum operations, respectively.
After fusion, non-maximum suppression is applied, and the top-k responses are selected as keypoints.
Both approaches consistently yield inferior performance across all keypoint budgets.
We further evaluate learnable routing variants, including a global router that takes texture features as input and predicts $M$ weights to perform image-level routing, 
and a soft router that uses patch-level routing probabilities to compute a weighted sum of detection maps instead of performing hard selection.
These results indicate that the proposed router more effectively selects suitable detectors across different regions.

\textbf{Reward Formulation.}  
The effect of the reward formulation is examined via the trade-off parameter $\alpha$, which balances the quantity and precision. 
Using either term alone yields suboptimal performance across most budgets. 
A balanced setting ($\alpha=0.5$) improves over single-term variants. 
The best performance is achieved by the full model with $\alpha=0.8$. 
This indicates that emphasizing keypoint quantity while accounting for precision yields more effective routing decisions.

\subsection{Efficiency Analysis}
\label{efficiency}
\begin{wraptable}{r}{0.42\linewidth}
    \vspace{-1em}
    \setlength{\tabcolsep}{3pt}
    \centering
    \caption{\textbf{Efficiency Analysis.}}
    \begin{tabular}{l ccccc}
    \toprule
        Method & Param. (M) & Time (ms) \\
        \midrule
        SP & 0.9 & 16.3 \\
        ALIKED & 0.4 & 17.2  \\
        DeDoDe & 14.6 & 65.8  \\
        DaD & 6.5 & 27.8  \\
        DeepLSD & 8.6 & 45.7  \\
        MuGE & 72.5 & 58.7  \\
        MoD & 9.4 & 37.4  \\
         \bottomrule
    \end{tabular}
    \label{tab:efficiency}
    \vspace{-4em}
\end{wraptable}

The efficiency comparison is reported in Table \ref{tab:efficiency}.
MoD maintains practical inference time while producing diverse detection maps and adaptive routing,
achieving efficient mixture of detectors without running all candidate detectors separately.

\subsection{Understanding MoD}
To gain deeper insights into the behavior of MoD, we visualize and analyze its key components.

\textbf{Budget-aware bias.}  
We visualize the learned budget biases for different detectors in Figure \ref{fig:bias}. 
Each row corresponds to a keypoint budget, and each column represents a detector. 
The learned biases exhibit clear and interpretable patterns across different budgets. 
At lower budgets, detectors such as DaD receive strong positive biases, 
while others are suppressed, indicating a preference for methods that produce sparse and reliable keypoints. 
As the budget increases, the preference gradually shifts toward detectors such as MuGE, 
suggesting that edge-based detectors become more advantageous when more keypoints are allowed. 
At the highest budget, the preference changes noticeably, where detectors such as DeDoDe and SuperPoint receive positive biases, 
while MuGE is strongly suppressed, reflecting different behaviors under dense keypoint regimes.

\textbf{Reward.}  
We visualize the reward distributions in Figure \ref{fig:router spv}, where brighter regions indicate higher rewards. 
The reward maps exhibit clear spatial structure, revealing which detector performs best at each location. 
DaD and DeDoDe are favored in richly textured regions, 
while line- and edge-based detectors dominate along structural boundaries. 
ALIKED and SuperPoint produce more balanced responses in moderately textured areas. 
Different detectors dominate in different regions of the same image, indicating that no single detector is universally optimal. 

\begin{figure*}
    \centering
    \includegraphics[width=\linewidth]{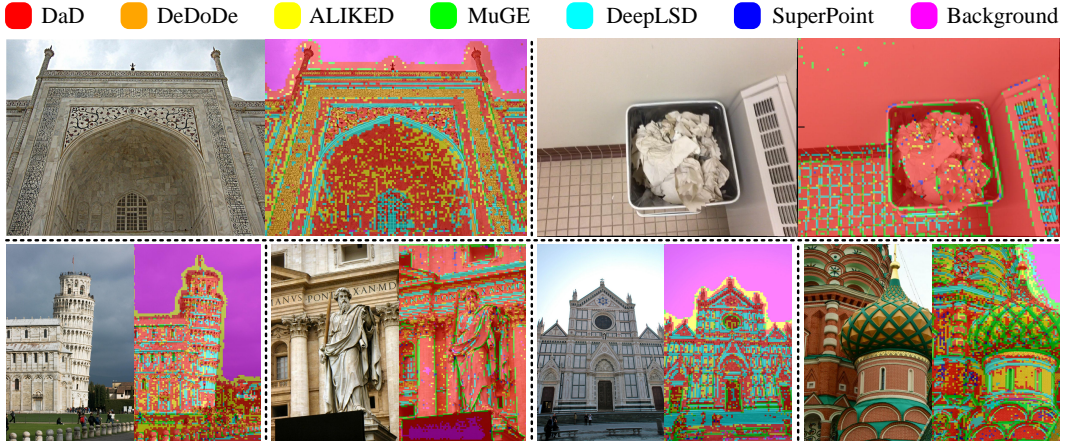}
    \caption{\textbf{Distribution of predicted routing decisions} (best viewed when zoomed in).}
    \label{fig:router pred}
\end{figure*}

\textbf{Routing decisions.}  
The predicted routing decisions are shown in Figure \ref{fig:router pred}, where each color corresponds to a selected detector or the background class. 
The routing maps exhibit strong spatial coherence, with detector assignments varying systematically across different regions. 
Point-based detectors are predominantly selected in textured areas, whereas line- and edge-based detectors are activated along prominent geometric structures. 
Regions with insufficient texture or unreliable geometry are assigned to the background class, avoiding ineffective keypoints.
Additional visualizations of routing decisions are provided in Appendix \ref{sec:pred_router} and \ref{sec:pred_router_num}.

\section{Limitations and Discussion}
\label{sec:discussion}
We investigate the use of pose estimation error as an additional reward signal for router training. 
However, this strategy does not yield observable improvements over the current design. 
This observation suggests that directly using downstream task errors as supervision is challenging, 
likely due to the indirect and noisy relationship between detector selection and final task performance. 
Future work may explore more principled approaches for incorporating downstream objectives, 
such as homography or pose estimation accuracy, into the reward formulation through more stable and structured supervision mechanisms.

\section{Conclusion}
\label{sec:conclusion}
We presented MoD, a unified framework that integrates multiple heterogeneous keypoint detectors through knowledge distillation and budget-aware routing. 
By distilling diverse detectors into a shared encoder with multiple lightweight heads,
MoD preserves their complementary inductive biases while achieving efficient inference within a single forward pass. 
Building on this, the proposed router adaptively selects detectors at the patch level by jointly considering local texture characteristics and global keypoint budget, 
and is trained via a reward signal that captures both keypoint quantity and precision.
Extensive experiments on MegaDepth, ScanNet, and HPatches demonstrate that MoD consistently outperforms strong baselines across a wide range of keypoint budgets, 
validating the effectiveness of both the distillation and routing components.

{
    \small
    \bibliographystyle{ieeenat_fullname}
    \bibliography{main}
}

\clearpage
\appendix

\section{Evaluation with RCM+}
\label{sec:rcm}

\begin{figure*}
    \centering
    \includegraphics[width=\linewidth]{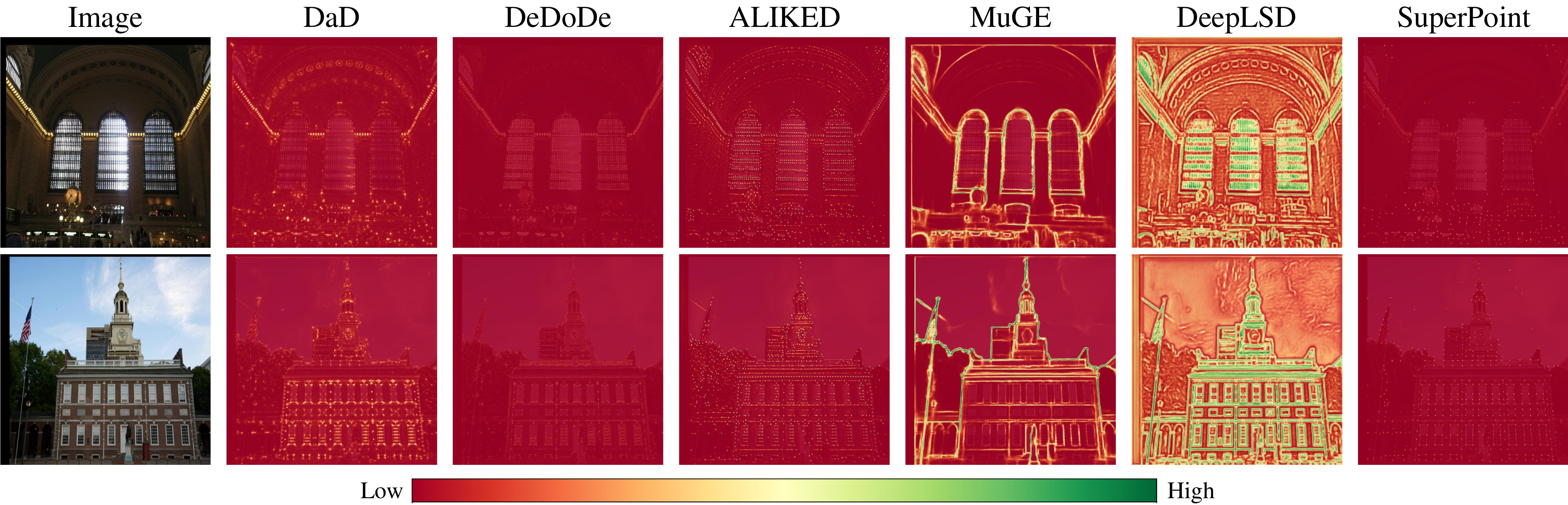}
    \caption{\textbf{The detection probability maps predicted by the distilled detection head} (best viewed when zoomed in).}
    \label{fig:scoremap}
    \vspace{-1em}
\end{figure*}

\begin{wraptable}{r}{0.55\linewidth}
    \vspace{-1em}
    \centering
    \caption{\textbf{Homography Estimation on HPatches using RCM+ \cite{rcm-plus} as the matcher.} 
    Performance is measured as the AUC@3 pixels of the end-point error.}
    \begin{tabular}{l ccccc}
    \toprule
        \multirow{2}{*}{Method} & \multicolumn{5}{c}{Homography Estimation AUC $\uparrow$} \\
        \cmidrule(lr){2-6}
        & 512 & 1024 & 2048 & 4096 & 8192 \\
        \midrule
        DeepLSD & 43.9	& 47.9	& 49.8	& 50.6	& 51.1 \\
        Canny & 34.1	& 40.2	& 46.8	& 50.7	& 51.3 \\
        UAED & 41.2	& 45.4	& 48.1	& 49.7	& 51.5 \\
        MuGE & 42.7	& 46.1	& 49.3	& 50.5	& 51.9 \\
        SIFT & 47.4	& 50.0	& 52.3	& 52.0	& 52.0 \\
        SP & 48.9	& 51.1	& 51.7	& 52.0	& 51.6 \\
        DISK & 41.0 & 44.2 & 46.4 & 47.8 & 49.0 \\
        ALIKED & 44.8	& 46.9	& 47.8	& 48.9	& 49.0 \\
        DeDoDe & 49.4 & 50.5 & 51.6 & 52.4 & 52.5 \\
        DaD & 49.9 & 51.2 & 52.0 & 51.8 & 51.6 \\
        MoD & \bf 50.4 & \bf 51.8 & \bf 52.2 & \bf 53.1 & \bf 53.3 \\
         \bottomrule
    \end{tabular}
    \label{tab:hpatches_rcm}
    \vspace{-2em}
\end{wraptable}

To further assess generalization across different matching pipelines, we additionally evaluate all methods using RCM+ as the matcher, as shown in Table \ref{tab:hpatches_rcm}.
MoD consistently achieves the best performance across all keypoint budgets.
The performance gains over strong baselines such as DaD and DeDoDe remain consistent, indicating that the advantages of MoD are not tied to a specific matcher.
Notably, the relative trends among detectors are largely preserved, 
further confirming that the proposed routing mechanism captures intrinsic detector complementarity rather than overfitting to a particular matching strategy.

\section{Distilled Detection Maps}  
We visualize the detection probability maps produced by the distilled heads in Figure \ref{fig:scoremap}. 
Each head exhibits a distinct response pattern that closely matches its corresponding teacher. 
In particular, point-based detectors produce sparse and localized activations around salient keypoints, 
whereas line- and edge-oriented detectors (e.g., DeepLSD and MuGE) strongly emphasize geometric structures such as edges and contours. 
These results demonstrate that the distillation process effectively preserves the heterogeneous inductive biases of different detectors within a unified architecture. 
Consequently, the distilled heads maintain complementary behaviors, providing a diverse set of candidates for subsequent selection.

\begin{figure*}
    \centering
    \includegraphics[width=\linewidth]{figures/router_pred_supp.pdf}
    \caption{\textbf{Distribution of predicted routing decisions} (best viewed when zoomed in).}
    \label{fig:router_pred_supp}
\end{figure*}

\section{Distribution of predicted routing decisions} 
\label{sec:pred_router} 
We visualize the distribution of predicted routing decisions under $2048$ keypoint budget in Figure \ref{fig:router_pred_supp}. 
Each image illustrates how the patch-wise router assigns detectors across diverse scenes, with colors representing different detectors and the background class. 
The results reveal that the router effectively adapts to varying local textures and scene structures, consistently assigning detectors to regions where they are most effective. 
Notably, the allocation patterns reflect detector complementarity: 
some detectors dominate structured or repetitive textures, while others are preferentially selected for edges, corners, or high-detail regions.
Background regions are largely assigned to the background class, indicating that the router successfully avoids assigning detectors to unstable or uninformative areas. 
These visualizations confirm that the routing decisions are spatially structured and align with the complementary strengths of the underlying detectors, 
providing interpretability for the learned adaptive allocation mechanism.

\begin{figure*}
    \centering
    \includegraphics[width=\linewidth]{figures/router_pred_num.pdf}
    \caption{\textbf{Routing decisions under varying keypoint budgets} (best viewed when zoomed in).}
    \label{fig:router pred num}
\end{figure*}

\section{Routing decisions under varying keypoint budgets}  
\label{sec:pred_router_num} 
We further analyze the effect of keypoint budget in Figure \ref{fig:router pred num}.
As the budget increases, the routing patterns change noticeably. 
At lower budgets, the router favors detectors that produce sparse and reliable keypoints such as ALIKED and DaD. 
In contrast, at higher budgets, router increases activation of detectors that can generate a large number of valid keypoints in highly textured regions, such as DeDoDe. 
This behavior reflects the model's ability to adapt detector selection according to the desired keypoint density.

\begin{figure*}
    \centering
    \includegraphics[width=\linewidth]{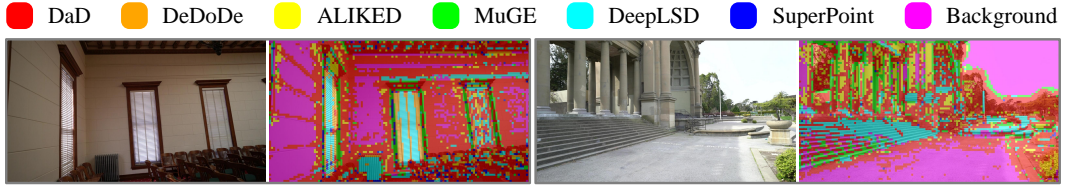}
    \caption{\textbf{Failure cases of the router} (best viewed when zoomed in).}
    \label{fig:failure cases}
\end{figure*}

\section{Failure Cases}  
While MoD generally produces robust keypoint distributions, some texture-less regions, such as floors or walls, 
are occasionally classified as background despite containing useful keypoints. 
Figure \ref{fig:failure cases} shows examples of such misclassifications. 
Analyzing these cases highlights limitations of the current reward and routing mechanisms and suggests potential improvements, 
such as incorporating additional context or uncertainty-aware routing.

\begin{figure*}
    \centering
    \includegraphics[width=\linewidth]{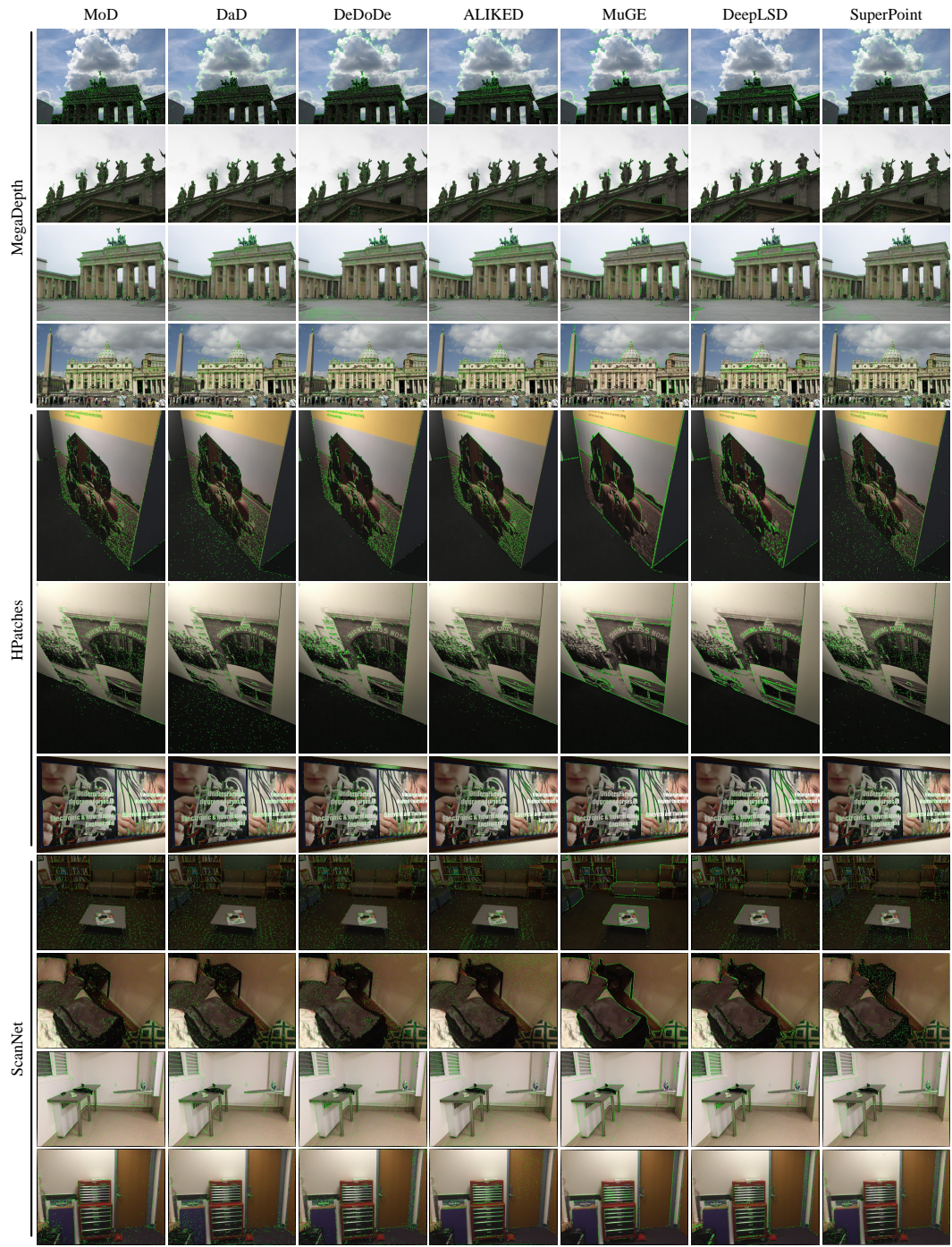}
    \caption{\textbf{Keypoint visualization} (best viewed when zoomed in).}
    \label{fig:kpt_viz}
\end{figure*}

\section{Keypoint Distribution}  
We visualize the final keypoint distributions predicted by the seven detectors on MegaDepth, ScanNet, and HPatches, 
with a unified keypoint budget of $2048$, as shown in Figure \ref{fig:kpt_viz}. 
The visualizations reveal that MoD produces a more balanced and spatially diverse distribution compared to individual detectors. 
While single detectors tend to concentrate keypoints in specific texture patterns or edges, MoD effectively integrates the complementary strengths of all detectors, 
covering both high-texture and low-texture regions across scenes. 
This results in a more uniform keypoint coverage, which aligns with the adaptive routing decisions and contributes to improved downstream geometric performance. 
The figures also highlight that detectors specialize in different scene components, 
reinforcing the importance of detector complementarity in generating robust keypoint distributions.


\clearpage

\end{document}